\documentclass{article}

\usepackage{float}
\usepackage{arxiv}

\usepackage[utf8]{inputenc} 
\usepackage[T1]{fontenc}    
\usepackage{hyperref}       
\usepackage{url}            
\usepackage{booktabs}       
\usepackage{amsfonts}       
\usepackage{nicefrac}       
\usepackage{microtype}      
\usepackage{lipsum}
\usepackage{graphicx}
\usepackage{multirow}
\usepackage{color, colortbl}
\usepackage{authblk}
\definecolor{lightgray}{rgb}{0.83, 0.83, 0.83}
\graphicspath{ {./images/} }

\title{Spanish Biomedical Crawled Corpus: A Large, Diverse Dataset for Spanish Biomedical Language Models}

\author[1]{
 Casimiro Pio Carrino
}
\author[1]{Jordi Armengol-Estap\'e
}
\author[1]{Ona de Gibert Bonet
}
\author[1]{Asier Gutiérrez-Fandiño
}
\author[1]{Aitor Gonzalez-Agirre
}
\author[1]{Martin Krallinger
}
\author[1]{Marta Villegas
}
\affil[1]{Text Mining Unit\\
Barcelona Supercomputing Center
}
\affil[ ]{\textit {\{casimiro.carrino,jordi.armengol,ona.degibert,asier.gutierrez,aitor.gonzalez,martin.krallinger,marta.villegas\}@bsc.es}}

\begin{document}
\maketitle
\begin{abstract}
We introduce CoWeSe (the Corpus Web Salud Español), the largest Spanish biomedical corpus to date, consisting of 4.5GB (about 750M tokens) of clean plain text. CoWeSe is the result of a massive crawler on 3000 Spanish domains executed in 2020. The corpus is openly available and already preprocessed. CoWeSe is an important resource for biomedical and health NLP in Spanish and has already been employed to train domain-specific language models and to produce word embbedings. We released the CoWeSe corpus under a Creative Commons Attribution 4.0 International license in Zenodo (\url{https://doi.org/10.5281/zenodo.4561970}).

\end{abstract}

\section{Introduction and Motivations}
Transfer learning has revolutionized virtually all tasks in the Natural Language Processing field \cite{ruder-etal-2019-transfer}. It turned out to be exceptionally successful when large-scale language models leveraging unlabelled data to perform self-supervised learning were employed. Two paradigmatic examples are the GPT \cite{Radford2018ImprovingLU}, and BERT \cite{devlin-etal-2019-bert}. However, although gathering unlabelled data (raw text) is considerably cheaper than producing annotations, obtaining high-quality text is especially challenging in the biomedical and health domains for non-English languages. 

Following the paradigm of "the web as a corpus", manually crawling websites belonging to the target domains of interest is a strategy worth exploring. The CommonCrawl\footnote{\url{https://commoncrawl.org/}} is very a large repository of crawled websites. However, it needs preprocessing to extract the text relevant to the user. The OSCAR corpus \cite{DBLP:journals/corr/abs-2006-06202} was built applying language identification to CommonCrawl. In the case of the biomedical domain in English, BioBERT \cite{10.1093/bioinformatics/btz682}, and SciBERT \cite{beltagy-etal-2019-scibert} aggregated and processed different corpora (mostly proceeding from articles) to develop domain-specific BERT models. In the case of biomedical data in Spanish, there is an ongoing effort to develop textual resources for the biomedical domain \cite{gonzalez-agirre-etal-2019-pharmaconer, miranda2020named, soares-etal-2019-medical}. However, there are still few corpora available. Furthermore, a comprehensive biomedical corpus should ideally allow the transfer to the even more low-resourced sub-domains, such as the clinical one.
Therefore, the CoWeSe represents an unprecedented effort to build the largest biomedical and health-related corpus in Spanish to the best of our knowledge. Unlike other works, we crawl the web instead of scientific literature, providing a large-scale corpus with diverse contents.

The remainder of this paper is organized as follows. First we describe the data collection process in Section \ref{sec:collection}, and the preprocessing strategy in Section \ref{sec:preprocessing} providing some basic statistics of the corpus. Finally, in Section \ref{sec:con} we conclude with some final observations.

\section{Data Collection}
\label{sec:collection}
We crawled the web using 3,338 manually curated links as seeds with a depth of 5. They were selected to include diverse and relevant content, including websites categorized as Sites of Interest for Health by the Carlos III Health Institute (ISCIII).\footnote{\url{https://www.isciii.es/QueHacemos/Servicios/Biblioteca/Paginas/default.aspx}} Although the majority of the content is in Spanish; content in Catalan, Galician and Basque have been also included.

The crawling was performed during the first half of 2020, and we exclusively scraped websites whose robots files allowed it, resulting in a total of 2,766 websites. The raw crawling size is about 905GB of WARC files. When extracting the text, we only considered the paragraph and headers HTML tags. Formats other than HTML were not included.  

The selected websites belong to at least one of the following categories: \begin{enumerate}
        \item Medical communities.
        \item Scientific communities.
        \item Medical journals.
        \item Research centres.
        \item Pharmaceutical companies.
        \item Informative websites about health issues.
        \item Patient associations.
        \item Personal blogs from healthcare professionals.
        \item Hospital websites.
        \item Public health organizations.
    \end{enumerate}

\section{Preprocessing}
\label{sec:preprocessing}
We use the cleaning pipeline\footnote{\url{https://github.com/TeMU-BSC/corpus-cleaner-acl}} introduced in \cite{armengol-estape-etal-2021-multilingual} that is a series of customized components performing data parsing in different formats, sentence splitting, language detection, removal of noisy and ill-formed sentences, content deduplication and eventually output the data with their original document boundaries. Due to the vast amount of data, we deployed the cleaning pipeline across 50 nodes of a High-Performance Computing cluster\footnote{\url{https://www.bsc.es/support/MareNostrum4-ug.pdf}}.

The pipeline is parametrized to allow configuring its action according to the nature of the data to be processed. We then modify the cleaning pipeline's parameters to adapt the components to peculiarities of the biomedical and health domains. Specifically, we increased the default language identification thresholds since we found that biomedical Spanish is often predicted as Spanish with a lower probability than general domain Spanish. We also decreased the allowed minimum length for each document since the overall crawling is not big enough to be too aggressive in this regard.
After preprocessing, we obtained a cleaned corpus of 4.5GB of plain text from the original 905GB of WARC files. Table \ref{tab:stats} shows some statistics of the cleaned corpus. 
\begin{figure}[H]
\centering
\begin{tabular}{l|r}
\toprule
Size (plain text) & 4.5GB \\
Tokens (\texttt{wc} word count) & 745.70M \\
Documents & 1.58M \\ 
Sentences & 32.77M \\
\bottomrule
\end{tabular}
\caption{\label{tab:stats}
Basic statistics of the clean corpus.
}
\end{figure}

\section{Access}
\label{sec:acc}
The CoWeSe corpus is openly available under a Creative Commons Attribution 4.0 International license in Zenodo\footnote{\url{https://doi.org/10.5281/zenodo.4561970}}.

\section{Conclusions}
\label{sec:con}

In this work, we have introduced the CoWeSe corpus, the largest Spanish biomedical corpus to date. From one side, the data collection process ensures vast size while gathering diverse content. On the other side, the cleaning preprocessing produced high-quality and easy to use textual data. We believe the CoWeSe corpus will have a significant impact for the biomedical NLP community encouraging the development of biomedical and health-related language models and tools in Spanish.

\section*{Acknowledgements}
This work was partially funded by the Spanish State Secretariat for Digitalization and Artificial Intelligence (SEDIA) within the framework of the Plan-TL, the Future of Computing Center, a Barcelona Supercomputing Center and IBM initiative (2020).

\bibliographystyle{plain} 
\bibliography{references
} 
\end{document}